\documentclass[letterpaper]{article} 
\usepackage{aaai20}  

\usepackage{times}
\usepackage{epsfig}
\usepackage{graphicx}
\usepackage{amsmath}
\usepackage{amssymb}

\usepackage[utf8]{inputenc}
\usepackage[english]{babel}

\usepackage[nottoc]{tocbibind} 

\begin{document}

\title{Learning to Conceal: A Deep Learning Based Method for Preserving Privacy and Avoiding Prejudice}

\author{Moshe	Hanukoglu, Nissan	Goldberg, Aviv	Rovshitz, Amos Azaria \\
Computer Science Department, Ariel University, Israel}

\maketitle

\begin{abstract}

In this paper, we introduce a learning model able to conceals personal information (e.g. gender, age, ethnicity, etc.) from an image, while maintaining any additional information present in the image (e.g. smile, hair-style, brightness). Our trained model is not provided the information that it is concealing, and does not try learning it either. Namely, we created a variational autoencoder (VAE) model that is trained on a dataset including labels of the information one would like to conceal (e.g. gender, ethnicity, age). These labels are directly added to the VAE's sampled latent vector. Due to the limited number of neurons in the latent vector and its appended noise, the VAE  avoids learning any relation between the given images and the given labels, as those are given directly. Therefore, the encoded image lacks any of the information one wishes to conceal. The encoding may be decoded back into an image according to any provided properties (e.g. a 40 year old woman).
%

The proposed architecture can be used as a mean for privacy preserving and can serve as an input to systems, which will become unbiased and not suffer from prejudice.
We believe that privacy and discrimination are two of the most important aspects in which the community should try and develop methods to prevent misuse of technological advances.

\end{abstract}

\section{Introduction}
On April 10, 2018, 
following the Cambridge Analytica scandal and a data leakage of more than 87 million users,
Facebook Founder and CEO, Mark Zuckerberg, was summoned to testify before the Senate Judiciary and Commerce Committees about steps he intends to take to ensure Facebook users’ information will not be abused again. Zuckerberg was also required to explain how he intended to prevent third parties from using the platform inappropriately \cite{confessore2018cambridge}.

There are many implications of user privacy with respect to user data; foremost is the fear of exposing personal information over a social network. As indicated by Almadhoun et al. \shortcite{almadhoun2011perceived} 75.8\% of the respondents do not believe that they would feel totally safe when providing sensitive information about themselves over the social networks. Indeed, several network attacks exist that allow strangers to extract personal information from a victim \cite{zhou2008preserving}. Therefore, any personal data uploaded to the internet might be exposed by a third party who should not be permitted to view it. Data anonymization methods could support user privacy.

In addition, user-privacy also relates to research communities. Rothstein states that the current regulatory frameworks of the Common Rule and Privacy Rule emphasize privacy interests, but they overlook the privacy interests of individuals whose health information and biological specimens are used in research without their knowledge, consent, or authorization \cite{rothstein2010deidentification}. Methods for data anonymization would allow faster and more effective research, including life-saving and medical research, as people are more likely to be willing to share truly anonymized data.



\begin{figure}
  \centering
  \includegraphics[width=0.5\textwidth]{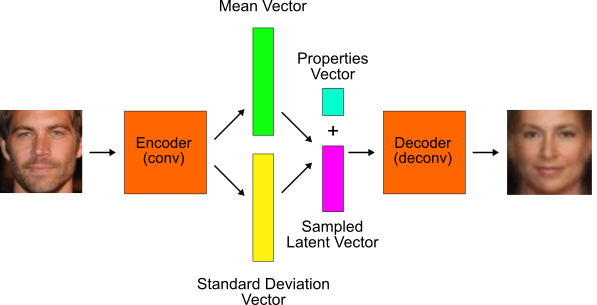}
  \caption{General architecture of The Blind Autonencoder for Fairness and Objectiveness (BAFO).}  
  \label{fig:general}
\end{figure}

Another important problem is the issue of discrimination. 
According to an Analysis commissioned by the British daily newspaper The Guardian, information from the London Mayor’s Office for Policing and Crime, shows that while black people compose only 15.6\% of London’s population (and white people compose 59.8\% of them), in 2018, 43\% of searches were of black people, while only 35.5\% were of white people \cite{black2019}. Furthermore, searches of black people were less likely to detect crime than those conducted on white people. As obtained from the data, while 21\% of searches of white people led to an arrest, only 16\% of searchs on black people led to an arrest. In addition, it seems that ``disproportionality has increased'', with the likelihood of black people being stopped being 4.3 times higher than white people in 2018, compared with 2.6 times more likely in 2014 \cite{black2018}. 
Unfortunately, it has been shown that computer programs and machine learning algorithms suffer from prejudice and gender bias as well \cite{leavy2018gender,vincent2018google}.

To overcome the issues of privacy and prejudice in images, we introduce the Blind Autonencoder For Fairness and Objectiveness (BAFO), a novel deep learning architecture that is based on a variation autoencoder (VAE) \cite{pu2016variational}. Namely, BAFO is trained on a dataset including labels of the information one would like to conceal (e.g. gender, ethnicity, age). These labels are directly added to the VAE's sampled latent vector (see Figure \ref{fig:general}). 
Due to the limited number of neurons in the latent vector and its appended noise, the VAE avoids learning any relation between the given images and the given labels, as those are given directly. Therefore, the encoded image lacks any of the information one wishes to conceal; BAFO is practically blind to all this information. The encoding may be decoded back into an image according to any provided properties (e.g. a 40 year old woman).

It might seem unintuitive that during training BAFO is given the information that it should later conceal. However, BAFO's behavior might be similar to a child that is asked to use a calculator from the very beginning of preschool. The child might focus on acquiring complex mathematical skills but is not likely to know how to add or multiply numbers by herself, because she learns to trust the calculator instead. Furthermore, if that child will later use a different calculator that uses different functions, the computation results will become different. Another similar example is the autopilot, which may cause pilots to not be able to fly an airplane themselves, because they learn to trust the autopilot \cite{carr2014glass}.

Using BAFO, security offices may monitor concealed surveillance footage, that is, surveillance footage in which all information required to be concealed (e.g. gender, race) is not present. It is important to note that this information is not only removed, but explicitly not learned by BAFO. Furthermore, such footage may be fed into an artificial intelligence system that will detect suspicious act, but will be totally unbiased, as the footage will not include any of the concealed information, or even any information that will allow the system to deduct the concealed information from.

The general idea behind BAFO is to explicitly provide a learner, during its learning process, with the information that it should conceal during the inference phase, in which this information will not be provided. We believe that this idea is not limited to images and videos, but is more general and can easily adapted to be used for concealing one's voice, and text. Concealing voice may be used by the press or by court when the identity of the speaker needs to remain unknown. Concealed text may be used when applying for education or job openings. A candidate may fear to fall for prejudice, and may therefore wish to hide her gender or race from the curriculum vitae (CV), in a way that this information cannot be deduced.  
Text concealing can be used also for robustly annonymizing data. This is because BAFO would remove not only any explicit identifiers (e.g. name and country of birth), but also from any implicit ones (e.g. specific expressions used by some group of people). In addition, users knowing that their data is totally annonymized (either by a company or even by running a system simalar to BAFO themselves) are more likely to share their data. Data sharing is especially important in any medical related system, which may literally save people's life.

It is very important to note that the idea behind BAFO is very different than other tools that may be used to convert one type of image to another (e.g. showing an older version of one-self) \cite{akhtar2019face}. Such technologies apply a smart filter that converts some type of image to another, this filter is applied regardless of the original image. Therefore, in such systems if a feminine filter is applied on a woman, she would seem even more feminine. Unlike with BAFO, if trained to remove gender, in which all gender related aspects of any image are totally removed, and are later explicitly added in order to produce a new image. 

\section{Related work}

Several previous works have developed different machine learning architectures with an attempted to model the age progress in images \cite{ramanathan2006modeling,ramanathan2008modeling,zhang2017age,baek2018editable}.
Zhang et al. \shortcite{zhang2017age} developed a Conditional Adversarial AutoEncoder (CAAE) model that studies the facial features, and important parameters that appear in each age segment. When given an image it may be converted to a new image by progression or regression of the current age.
In the first stage the face is mapped to an invisible vector, encoded by conventional coding.
The hidden layer retains the characteristics of the facial features in a ``personal'' manner. There are two networks that improve each other and finally compare the result obtained with the real one. While Zhang et al. is, to the best of our knowledge, the only work that does not require to be given the age of a given image in order to output a specific age, their work is still very different than ours, as for training, their model requires many images of the same person at different ages in order. CAAE uses the dataset to learn how age progresses / regresses and creates an age manifold that can then be used to create a new image.

In recent years there have been a number of popular approaches for creating artificial images:
Generative Adversarial Network (GAN) \shortcite{goodfellow2014generative}
is a Generative model for creating new information such as creating fake high quality images. GANs are trained to output images that look real, and are therefore sharp and have high contrast.
GANs includes a discriminator and a generator; both components compete with each-other. 
The discriminator identifies whether each image is original or has been created by the generator, while the generator’s goal is to create images that will seem real to the discriminator. That is, the generator tries to deceive the discriminator into thinking that the the images created by the generator are original.
Baek et al. \shortcite{baek2018editable} created a face editing tool that is based on GANs.
However, GANs are not appropriate for our goal, since we do not intend to produce an image that looks are real as possible, but to preserve the original image wile concealing the properties we want to conceal.

Another deep learning based approach for generating images is the Variational AutoEncoder (VAE) \cite{doersch2016tutorial}. Similar to a standard AutoEncoder, a VAE architecture includes a bottleneck and, during training, it tries to restore a given image. The representation in the bottleneck is in fact a compressed representation of the given image. However, a VAE learns two vectors, a mean vector and a standard deviation vector, the latent vector is sampled using the mean and standard deviation vectors. 
VAEs have been used for denoising and generating images that are similar to the training set \cite{huang2018introvae}.
In this paper we propose a novel concept in which the VAE that is trained on the given data, but the labels are appended directly to the latent vector (see Figre \ref{fig:general}). The VAE therefore does not attempt to learn any information that is directly provided, and becomes blind to these properties. While not being the primary intention of our VAE architecture, we note that it can also be used to generate images with specific attributes (e.g. only images of a 40-year old female).



\section{Dataset}
We use the UTKFace dataset \cite{zhang2017age}, which contains approximately 23,000 headshot photos. The images in the dataset are labeled with the age (ages range from 0 to 116), gender (male and female) and ethnic origin which is divided into five types. 
The dataset was split to 85\% training-set and 15\% test-set.
All our software is available at:[LINK REMOVED FOR ANONYMITY OF THE SUBMISSION].



\begin{figure*}[!h]
\centering
\includegraphics[width=1\textwidth]{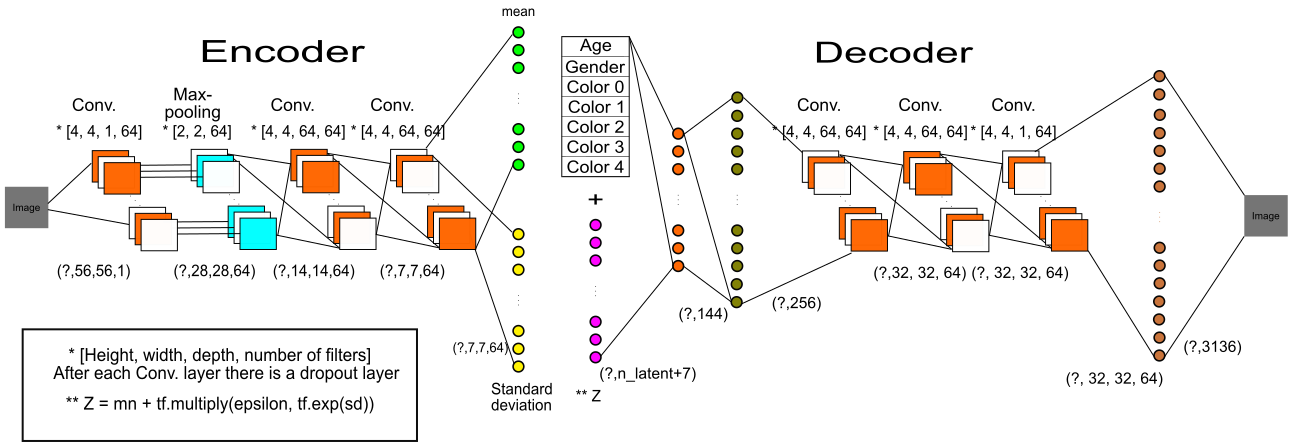}\hfill
\caption{The Architecture of the VAE.}
\label{fig:VAE}
\end{figure*} 

\section{Method}
In this section we introduce the Blind Autonencoder For Fairness and Objectiveness (BAFO). 

The general architecture of the model is depicted in Figure \ref{fig:VAE}.
In the training phase, a labeled photo is inserted as an input to BAFO. In the UTKFace dataset the photos size is 56*56*3 (RGB), and the photos are tagged by age, gender, and origin. We note that the labels include the information that we intend to later conceal.
The image is then compressed by the encoder into two vectors representing the parameters that describe the image, one vector for the mean and the other for the standard deviation. After sampling the latent vector from the mean and the standard deviation, it is concatenated to the the information BAFO is concealing, the age, the gender and the origin. The latent vector is then decoded back to an image with a size similar to the input image.

The motivation behind this architecture is that, since the information to be concealed is provided without noise, the system will be able to devote all of its efforts to learning only the additional parameters that affect the image, and not the information that is explicitly provided.

\subsubsection{Latent Vector Size}
We consider two different sizes for the latent vector, 48 and 100. In order to evaluate the performance of BAFO and determine which of vector size to use, we concealed the test data and decoded it into females and males in 5 age groups: 1 year old, 20 year old, 40 year old, 60 year old and 80 year old. The gender of the decoded images was processed by Microsoft’s Azure Face API, which, fortunately, correctly classified all images according to their intended gender. That is, all concealed male or female images that were decoded using a female value in the properties were correctly classified as females (similarly for males). This was true for both the architecture of BAFO with 48 and 100 latent vector sizes, which is quite a remarkable result.
In order to evaluate the decoded concealed images with respect to the age, we developed an age classifier. We note that the age classifier itself had a root mean squared error (RMSE) of 6.75.
The root mean squared error (RMSE) and mean absolute error (MAE) of the decoded concealed images appear in Table \ref{tbl:age}. As depicted by the table, BAFO with a latent vector of size 48 seems to slightly outperform BAFO with a latent vector of size 100.

\begin{table}
\begin{tabular}{l | c c | c c}
target &RMSE&RMSE&MAE&MAE\\ 
age & 48-cells & 100-cells & 48-cells & 100-cells \\ [0.5ex]
\hline
1    &\bf{12.62}	& 13.34	& 10.60	& 11.42  \\
20   &9.37	& \bf{9.00}	& 7.94	& \bf{7.70}   \\
40   &\bf{7.89}	& 7.97	& \bf{6.56}	& 6.65   \\
60   &\bf{18.03}	& 18.71	& \bf{16.45}	& 17.45  \\
80   &\bf{27.41}	& 28.30	& \bf{26.03}	& 27.26  \\
\hline
average & \bf{15.06}	& 15.47	& \bf{13.52}	& 14.10  \\

\end{tabular}
\caption{The root mean squared error (RMSE) and mean absolute error (MAE) of concealed images decoded as 1 year olds, 20, 40 60 and 80 year olds (lower is better).} 
\label{tbl:age}
\end{table}

\section{Results}
\begin{figure*}
\includegraphics[width=0.8\textwidth]{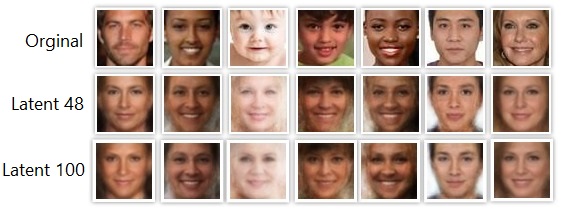}\hfill
 \centering
\caption{Images concealed by BAFO as a woman 40-year-old. In the first row are the input images; in the second row are the images concealed by BAFO when using a 48 length latent vector; and in the second row are the images concealed by BAFO when using a 100 length latent vector.
Note the way the system conceals only the age and gender of the given image, but preserves the smile, including the presence of teach, hair-style, light conditions, brightness of the images and the background.}
 \label{fig:collage_Woman_40_Latent_48_100}
\end{figure*}

Figure \ref{fig:collage_Woman_40_Latent_48_100} presents 7 randomly picked images from the test set, and the output of BAFO when concealed as a 40 year old woman, using a 48 and 100 length latent vector. In the first row are the input images; in the second row are the images concealed by BAFO when using a 48 length latent vector; and in the second row are the images concealed by BAFO when using a 100 length latent vector. Note that the facial features are smoother and delicate in the cheek and nose areas, as well as the thinner eyebrows in pictures 1,4 and 6, as shown in Figure \ref{fig:collage_Woman_40_Latent_48_100}. Note the way the system conceals only the age and gender of the given image, but preserves the smile, including the presence of teach, hair-style, light conditions, brightness of the images and the background.

\begin{figure*}
\includegraphics[width=0.8\textwidth]{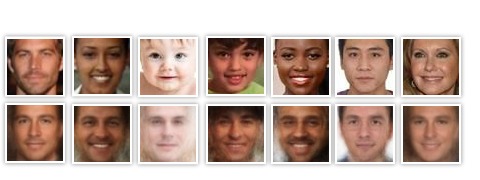}\hfill
 \centering
\caption{Concealing images as a 40-year-old male. In the first row are the input images and in the second row are the images concealed by BAFO when using a 48 length latent vector.}
\label{fig:collage_Man_40}
\end{figure*}
           
Figure \ref{fig:collage_Man_40} presents the same images when concealed as a 40 year old male. The male's facial features became more coarse in the center of the face, accompanied by wrinkles at the sides of her eyes and mouth, as can be seen in the second, fifth and seventh images. Note the smile, the length of the eyebrows and the structure of the nose are preserved in the second row  after the process is completed.

\begin{figure}
  \centering
  \includegraphics[width=0.46\textwidth]{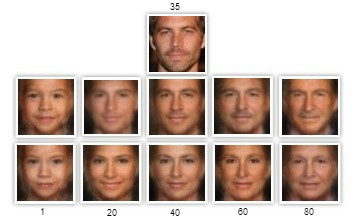}
  \caption{Concealing a single image as a man and woman at different ages. This may resemble common photo editor tools, however BAFO is very different than those tools as it is not rained to convert images from one demographic group to another, but is totally blind to the demographic group. The demographic group is explicitly \emph{added} to the image encoding in order to produce a meaningful image.}  
  \label{fig:collage_Paul_Walker_with_ages}
\end{figure}

In Figure \ref{fig:collage_Paul_Walker_with_ages}, we used a single image as the input and concealed it as a male and female with ages varying from a one year-old to an 80-year old person. 
As seen in the figure, BAFO can conceal any image at any range of ages and both as a male and a female. This result may resemble common photo editor tools (especially when making an image look older or younger) \cite{akhtar2019face}, which are mainly used for entertainment. However it is very important to note that BAFO is very different than those tools as it is not rained to convert images from one demographic group to another (or making an image older or younger), but is totally blind to the demographic group. Other photo editor tools are either directly trained to convert from one demographic group to another, or to add (and remove) specific features (such as aging features, feminine features, smile features etc.). 
Very importantly, BAFO is not provided the age, gender or race of the original image, nor does it learn it itself, as BAFO is not required to know whether the image should be converted to a younger image or an older image. Therefore, BAFO is truly unbiased and the demographic group is explicitly \emph{added} to the image encoding only in order to produce a meaningful image. 

\section{Discussion}
As depicted by Table \ref{tbl:age} and Figure \ref{fig:collage_Woman_40_Latent_48_100}, when modifying the latent vector size, there seems to be a trade-off between the image quality and the concealing performance. That is, with a large latent vector, the image decoded image quality is slightly better, but the image is slightly not concealed as well. This is expected, as the larger the latent vector, the more information BAFO can be stored in it, and the less does it need to rely on the additional information. On the other hand, the larger the latent vector, the more features may BAFO store in it and the higher the image quality. It is yet to be determine what the optimal latent vector size is, this may depend not only at the application, but also at the amount of labeled data available.  
Another approach is to increase the latent vector size, but also to modify the KL-divergence formula, so that the standard deviation will receive higher penalty rates, and therefore the data obtained from the mean-vector will be very noisy. This will further encourage BAFO to rely on the provided information, as much as possible.


\begin{figure*}
  \centering
  \includegraphics[width=0.7\textwidth]{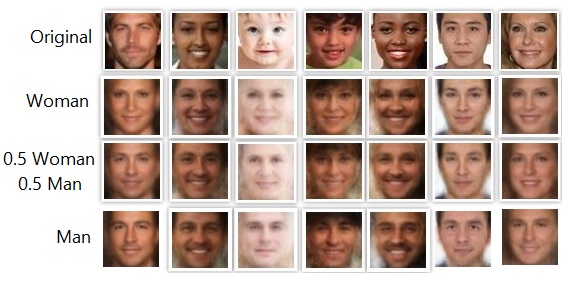}
  \caption{Concealing images as a 40-year-old male. In the first row are the input images and in the second row are the images concealed by BAFO when using a 48 length latent vector.}  
  \label{fig:half_woman}
\end{figure*}

In order to decode the concealed images, so that it is understandable for humans, FABO needs to be given some demographic information (or any other information related to the concealed information). However, it would be preferable if FABO could present the information using a neutral representation. For example, by using an image that is not gender associated. As a first approach to achieve such an image, we simply gave a value of 0.5 (the mean of the values for female and for male) as the gender value to FABO (See Figure \ref{fig:half_woman}). However, as can be seen in the figure this approach did not perform that well; this is not surprising, since no image with the value of 0.5 was given to FABO during the training phase. 
We note that finding a way to present the concealed images with a neutral gender or origin, etc., is a problem only relevant for concealed images that need to be viewed by humans, that is because if the concealed images are used by a machine, no decoding is required, and the machine will be unbiasedly trained and operated.

\section{Conclusions \& Future Work}
In this paper we introduce BAFO, an image concealer that receives as input an image and conceals unwanted properties (such as gender, ethnic origin, and age). The concealed images may be viewed by humans in a way that would remove any prejudice related to the concealed properties. Furthermore, these images can serve as input to another machine learning system, which, due to the input it receives from BAFO, will be unbiased. BAFO may also be used as a mean for privacy preserving by social network users. A user may conceal user private information in images (e.g. age, gender, ethnic origin) before she uploads them to a social network. Users who are familiar with that user and know the private information, will be able to decode the image according to the private information, and will view an image that is very similar to the original image that was uploaded. Other people, will see the images, but will not be able to extract the private information concealed in these images.

Extending the architecture described in this paper to concealing one's identity is very straight-forward. Such concealed images should preserve privacy and therefore could be used by researchers in different fields. Future work will also include the extension of BAFO's architecture to video, voice and text. Such extensions may have major implications on privacy preserving and unbiased systems, such as an \emph{unbiased} surveillance camera with automatic security threat detection. 







\clearpage
\bibliographystyle{aaai}
\bibliography{conceal}
\end{document}